\documentclass[letterpaper, 10 pt, conference]{ieeeconf}  

\IEEEoverridecommandlockouts                              

\usepackage{graphicx}  				
\usepackage{subfig}

\usepackage{cite}
\usepackage{balance}

\usepackage{standalone}
\usepackage{hyperref}

\hypersetup{
    colorlinks=true,
    filecolor=magenta,
    urlcolor=cyan,
}

\usepackage{amsmath}

\usepackage{tikz}
\usetikzlibrary{fit}
\usetikzlibrary{shapes,arrows}
\usetikzlibrary{calc}
\usetikzlibrary{shapes.multipart}
\usetikzlibrary{backgrounds}

\usepackage{booktabs}

\usepackage{float}
\usepackage{subfig}
\usepackage{xcolor}
\definecolor{rev1}{rgb}{0.36, 0.54, 0.66}
\definecolor{rev2}{rgb}{0.9, 0.17, 0.31}
\definecolor{rev3}{rgb}{0, 0.52, 0.30}

\graphicspath{{./figures/}}

\title{\LARGE \bf 
Vision-based system for a real-time detection and following of UAV}

\author{Antonella Barisic, Marko Car, Stjepan Bogdan
\thanks{Authors are with Faculty of Electrical and Computer Engineering,
University of Zagreb, 10000 Zagreb, Croatia
        {\tt\small \{antonella.barisic, marko.car, stjepan.bogdan\}@fer.hr}}
}

\begin{document}

\thispagestyle{empty}
\onecolumn

\textcopyright  2019 IEEE. Personal use of this material is permitted.  Permission from IEEE must be obtained for all other uses, in any current or future media, including reprinting/republishing this material for advertising or promotional purposes, creating new collective works, for resale or redistribution to servers or lists, or reuse of any copyrighted component of this work in other works.
\newline

Published in: 2019 Workshop on Research, Education and Development of Unmanned Aerial Systems (RED UAS)

DOI: 10.1109/REDUAS47371.2019.8999675

URL: \url{https://ieeexplore.ieee.org/abstract/document/8999675}
\url{}

\twocolumn

\maketitle

\thispagestyle{empty}
\pagestyle{empty}

\begin{abstract}

In this paper a vision-based system for detection, motion tracking and following of Unmanned Aerial Vehicle (UAV) with other UAV (follower) is presented. For detection of an airborne UAV we apply a convolutional neural network YOLO trained on a collected and processed dataset of 10,000 images. The trained network is capable of detecting various multirotor UAVs in indoor, outdoor and simulation environments. Furthermore, detection results are improved with Kalman filter which ensures steady and reliable information about position and velocity of a target UAV. Preserving the target UAV in the field of view (FOV) and at required distance is accomplished by a simple nonlinear controller based on visual servoing strategy. The proposed system achieves a real-time performance on Neural Compute Stick 2 with a speed of 20 frames per second (FPS) for the detection of an UAV. Justification and efficiency of the developed vision-based system are confirmed in Gazebo simulation experiment where the target UAV is executing a 3D trajectory in a shape of number eight.

\end{abstract}

\section{Introduction}
\label{sec:introduction}

Considering a thriving development of UAVs, the perception of their surrounding becomes essential. For this purpose, UAVs are usually equipped with visual sensors, radars or lasers. Due to the rapid development of computer vision techniques, vision-based applications of UAVs are becoming a promising solution for problems of surveillance \cite{surv}, navigation \cite{nav}, search and rescue \cite{s&r}, sense and avoid \cite{SAA}, etc. On the other hand, problem of detection and following of the object of interest can be a part of the solution or individual problem. The most common objects of interest in studies of detection and following by UAV are persons \cite{fallperson,Han2016DeepD} and vehicles \cite{veh_per,veh_yolo}. 

The ability of the UAV to autonomously detect and follow another UAV is utilised in military and security purposes, furthermore in various operations of cooperative robotics. There are many challenges like real-time constraints, cluttered scenes, false and missed detection, data association and object overlapping. Also, detection of the UAV is more complex as a result of disturbed motion and fast movement of propellers. To address these problems one needs to make a compromise between system accuracy and computational efficiency. 

In this paper, the proposed vision-based system is specifically designed to meet requirements of real-time execution on UAV. The system can be divided into three crucial components: object detection, moving object tracking and visual servoing of UAV. For object detection we apply convolutional neural network YOLO which has been trained on our custom dataset. Based on detection results, we conduct a state estimation of the object of interest by applying a discrete Kalman filter. The Kalman filter is used to enhance object position in case of missing detection and to smooth abrupt changes of a bounding box enclosing the object. Finally, we implement a visual servoing for UAV in order to follow the object of interest.

The paper is organised as follows: in section II we present object detection algorithm trained on custom dataset of UAV images. Section III describes implemented Kalman filter. Visual servoing strategy is depicted in section IV while experimental validation of the entire system is provided in section V.

\section{Object detection}
\label{sec:detection}

Object detection is a well-studied computer vision technique for simultaneous localisation and classification of distinct object instances in image or video frame. In the interest of achieving admirable speed while maintaining good accuracy, we apply a convolutional neural network YOLO (You Only Look Once) \cite{yolov1} for detection of UAVs. In contrast to a region-based detectors, YOLO frames an object detection as a regression problem and performs a prediction of a bounding box and class probabilities with a single network. Each bounding box enclosing the object can be described by ($x$,$y$) coordinates of its center, width and height ($w$,$h$) of the box, and the probability of a predicted class.

\begin{figure}[ht!]
    \centering
    \vspace*{0.1in}
    \includegraphics[width=0.48\textwidth]{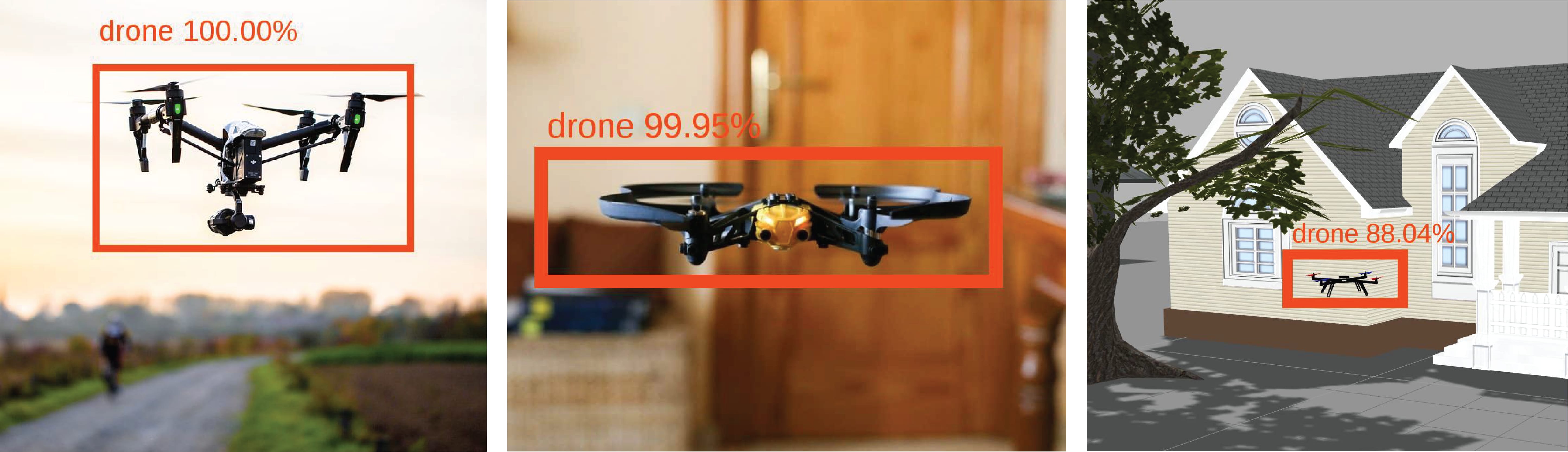}
    \caption{Example of detection results (bounding box and class confidence) of different UAVs in outdoor, indoor and simulation environment.}
    \label{fig:det}
\end{figure}

In this paper, YOLOv3 Tiny version is adopted, with a single class of UAVs to meet real-time constraints of entire vision-based system. A dataset of 10,000 UAV images was collected and processed for the purpose of this work. The dataset contains various types of UAVs in different surroundings. The network is trained on 70\% of images from the dataset and remaining images are used for validation presented in section V. The network we trained generalises well for both indoor, outdoor and even simulation environments (Fig.~\ref{fig:det}).

\section{Moving object tracking}
\label{sec:tracking}

\begin{figure}[ht!]
    \centering
    \vspace*{0.1in}
    \includegraphics[width=0.48\textwidth]{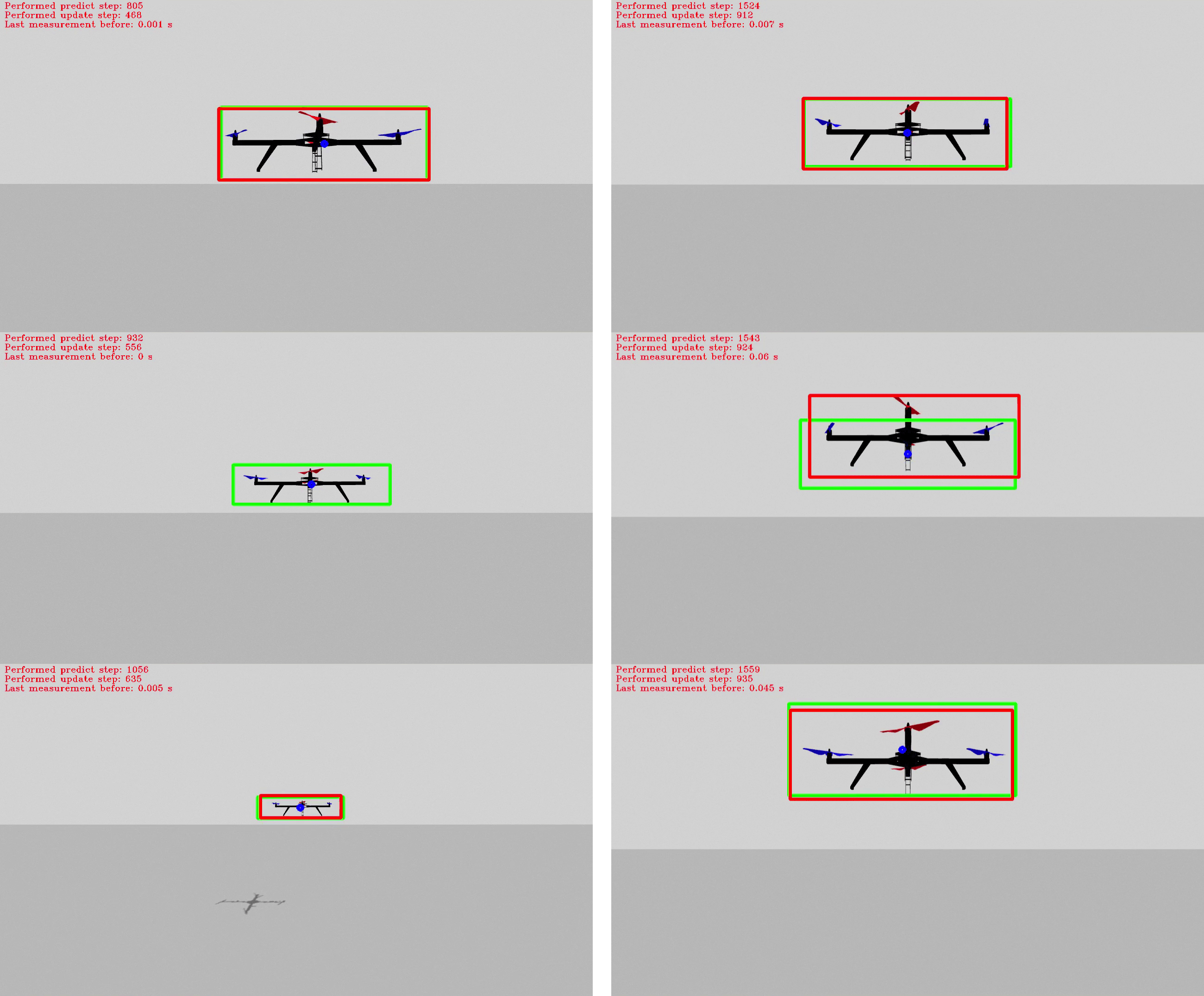}
    \caption{On the left side: sequence image of object motion represents object's departure from the follower UAV where Kalman filter predicts the state of the object in case of missing detection. On the right side: Kalman filter smooths abrupt changes of bounding box while the object is changing altitude. Green bounding box represents a Kalman filter estimation and red bounding box is an output of detection algorithm.}
    \label{fig:kalman}
\end{figure}

In the interest of following the target UAV, visual servoing strategy of the follower UAV must receive a continuous and steady visual information about object position. Detection results are sometimes too noisy which can badly affect performance of visual servoing. In case of UAV detection an additional alterations of bounding box due to the position of propellers in a captured frame has been encountered. The problem has been tackled by applying a discrete Kalman filter for estimation of dynamic states (position and velocity) of the target. The purpose of the Kalman filter is to prevent abrupt changes of the bounding box (Fig.~\ref{fig:kalman} right) and to predict object state in case of missing detection (Fig.~\ref{fig:kalman} left).

Equations of Kalman filter can be found in \cite{simon2006}. The state vector is given in the following form:

\begin{equation}
\label{eq:stanje_sustava}
    \hat{x}_k = \begin{bmatrix} xc_k & yc_k & w_k & h_k & 
    \dot{xc}_k & \dot{yc}_k & \dot{w}_k & \dot{h}_k \end{bmatrix} ^T
\end{equation}

where ($xc_k$,$yc_k$) are coordinates of the center of the bounding box, ($w_k$, $h_k$) are width and height of the bounding box and ($\dot{xc}_k$,$\dot{yc}_k$, $\dot{w}_k$, $\dot{h}_k$) are associated velocities. The  discrete  time  sample  is  denoted  by $k$. Kalman filter predicts the state based on a constant velocity motion model. Output of detection algorithm is used as a measurement in correction step of the filter. In case of multiple detection, the measurement association between the detected and tracked object is performed based on Intersection over Union (IoU). The process (Q) and measurement (R) noise are zero-mean Gaussian with the following covariance matrices:

\begin{equation}
    Q=
    \begin{bmatrix}
        4\cdot 10^{-5}\cdot I_4 & 0_4 \\
        0_4 & 0.4\cdot I_4
    \end{bmatrix},
        R=
    \begin{bmatrix}
        1 & 0 & 0 & 0 \\
        0 & 1 & 0 & 0 \\
        0 & 0 & 10 & 0 \\
        0 & 0 & 0 & 10
    \end{bmatrix}.
\end{equation}

\section{Visual servoing of UAV}
\label{sec:vs}

Visual servoing strategy has two main goals: to keep the object of interest in the field of view and to maintain object at a desirable distance. In order to fulfil these goals, a simple nonlinear controller has been devised. Keeping the object in the field of view is presented in a form of keeping its center near the center of image. Based on the Kalman filter estimation of $xc$ and $y_c$ coordinates,  translational errors in the camera frame have been converted to relative references for yaw and height of the follower UAV. Control commands for yaw and height are calculated as follows:

\begin{equation}
    \Delta \psi= \begin{cases}
    -\Delta \psi_{max} (2\cdot xc_{norm} -1), & \text{if }xc_{norm} > M\\
    \Delta \psi_{max} (1-2\cdot xc_{norm}),   & \text{if }xc_{norm} < N \\
    0,                            & \text{otherwise}
    \end{cases}
    \label{eq:yaw}
\end{equation}

\begin{equation}
    \Delta h= \begin{cases}
    -\Delta h_{max} (2\cdot yc_{norm} -1), & \text{if }yc_{norm} > M\\
    \Delta h_{max} (1-2\cdot yc_{norm}),   & \text{if }yc_{norm} < N \\
    0,                            & \text{otherwise}
    \end{cases}
    \label{eq:visina}
\end{equation}

where $\Delta \psi_{max}$ and $\Delta h_{max}$ are maximum values of references,
$xc_{norm}$ and $yc_{norm}$ are estimated coordinates of the object center relative to the image size, and $M$ and $N$ are controller constants implemented to avoid unnecessary activation of the visual servoing controller. By scaling a value of reference based on a deviation from the center, smooth motion has been achieved to avoid motion blur and react more accordingly to the motion of the object of interest.

Having the object in the field of view, the goal is to maintain approximately a constant area of the object in image plane, i.e. keep the object at a desirable distance. Therefor, a third control command for pitch angle has been generated:

\begin{equation}
    \Delta \theta= \begin{cases}
    \Delta \theta_{max} (\frac{d_{svr}-V}{d_{area,max}}), & \text{if }d_{svr} > V\\
    -\Delta \theta_{max} (1-\frac{d_{svr}}{W}),   & \text{if }d_{svr} < W \\
    0,                            & \text{otherwise}
    \end{cases}
    \label{eq:th}
\end{equation}

where $\Delta \theta_{max}$ is maximum value of pitch relative change, $d_{svr}$ is distance estimated by $\nu$-SVR model described next, $d_{area,max}$ is maximum distance at which change of object size is noticeable and $V$ and $W$ are controller constants.

\subsection{Distance estimation}
\label{subs:dist_est}

In order to estimate the distance between the follower UAV and the object of interest, a $\nu$-SVR (Support Vector Regression) model \cite{Scholkopf:2000:NSV:1139689.1139691} has been trained. Like all models from support vector family, SVR attempts to maximise the width of a margin for given training data. The trade-off between size of the margin and the penalty for errors is controlled by regularization parameter $C$. Support vectors are the data points that lie closest to the maximum margin plane. In $\nu$-SVR, the parameter $\nu$ is used to determine the number of support vectors compared to the total number of training data. The model is trained on 10,000 data acquired from simulation experiments described in the following section. Trained $\nu$-SVR takes [$xc$,$yc$,$w$,$h$] as input for distance estimation. Besides area of the object, a viewing angle is taken into account to achieve more accurate output. For training, we use a radial base function kernel defined by $\gamma$ parameter. Parameters of the model are optimised by employing a grid search and ten-fold cross-validation, producing a following values: $C=62,5$, $\nu=0,09$ and $\gamma=0,50625$. The mean squared error of distance estimation was $0.3133$ m.

\section{Experimental results}
\label{sec:results}

In this section, we evaluate performance of each implemented algorithm and describe experimental setup. As already mentioned, detection of UAV is performed on Intel Neural Compute Stick 2 (NCS2) using OpenVINO Toolkit specialised for convolutional neural networks like YOLO. In order to convert the model to an Intermediate Representation (IR) suitable for OpenVINO and to run inference on NCS2, original YOLO is transformed into Tensorflow representation. Trained model has been validated on 30\% of acquired dataset. Achieved results are presented in Table ~\ref{tab:valid}. Given the fact that the goal was to detect any kind of UAV in various surroundings, obtained results are very promising. 


\begin{table}[h!]
\centering
\caption{UAV detection results}
\begin{tabular}{r|c|c|c}
& \begin{tabular}{@{}c@{}}True \\ positives\end{tabular}
& \begin{tabular}{@{}c@{}}False \\ positives\end{tabular} 
& \begin{tabular}{@{}c@{}}False \\ negatives\end{tabular}\\ \hline \hline
Number of objects & 2866 & 205 & 646\\ \hline
Percentage & 75,06\% & 5,37\% & 16,91\% \\ \hline
\multicolumn{4}{c}{} \\  
& Precision & Recall & mAP \\ \hline \hline
Percentage & 93\% & 82\%  & 80,18\% \\
\end{tabular}
\label{tab:valid}
\end{table}

\begin{figure}[ht!]
    \centering
    \vspace*{0.1in}
    \includegraphics[width=0.48\textwidth]{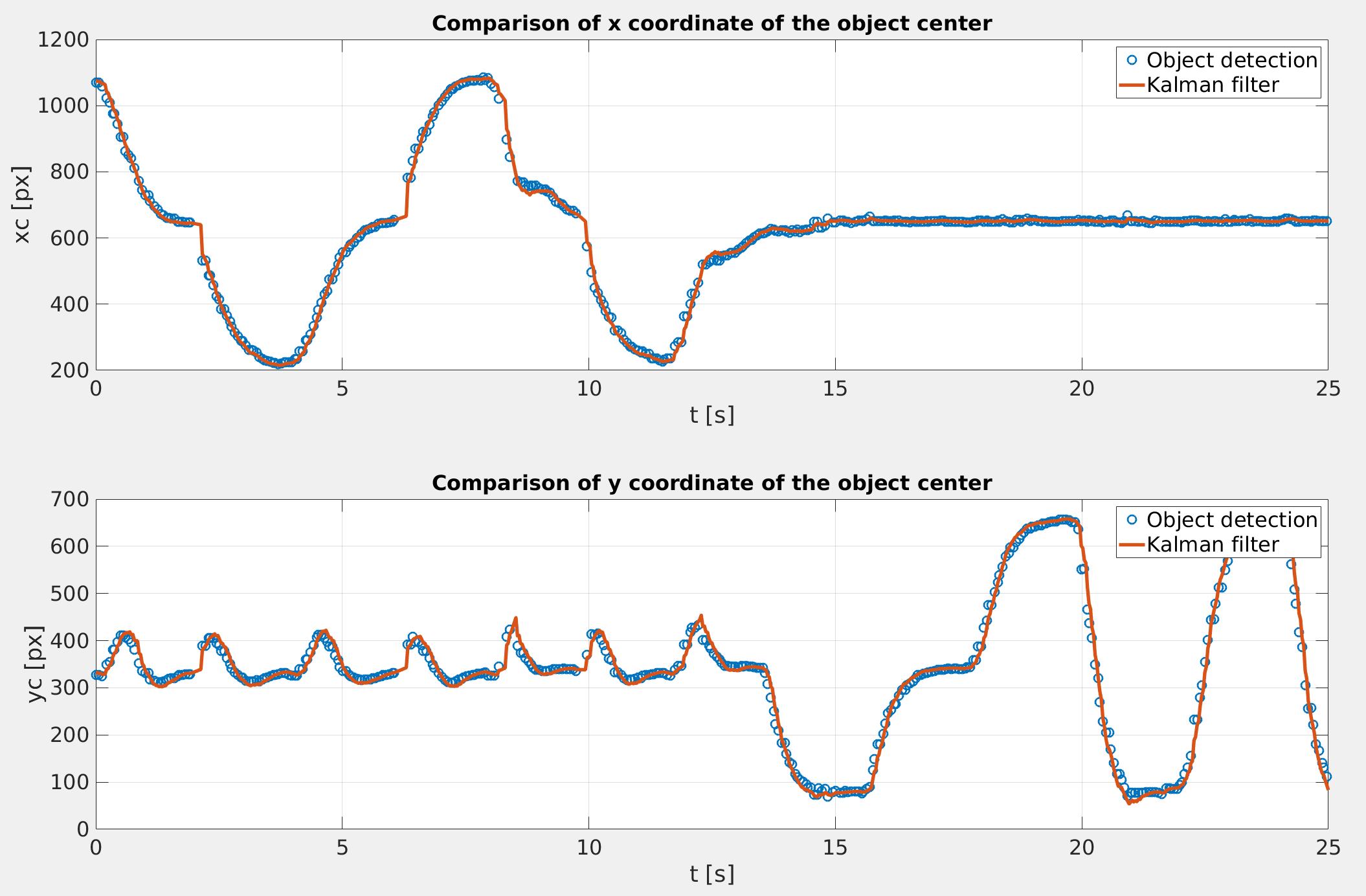}
    \caption{Comparison of Kalman filter estimation and object detection output of $xc$ and $yc$ coordinates of the bounding box center.}
    \label{fig:xy}
\end{figure}
\begin{figure}[ht!]
    \centering
    \vspace*{0.1in}
    \includegraphics[width=0.48\textwidth]{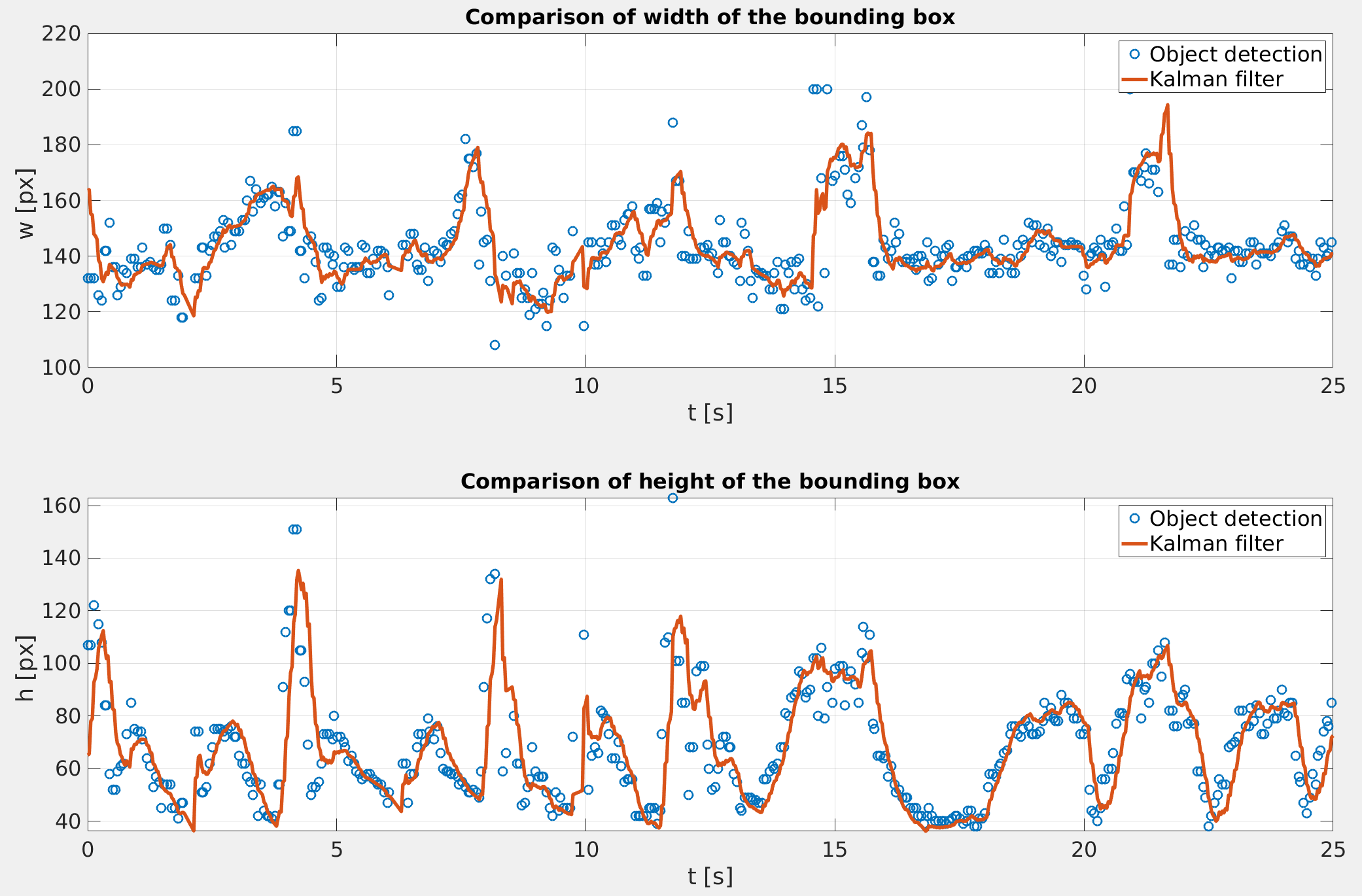}
    \caption{Comparison of Kalman filter estimation and object detection output of width and height of the bounding box.}
    \label{fig:wh}
\end{figure}

Simulation experiments in Gazebo have been performed in order to test tracking and visual servoing. The setup consists of two 3DR Arducopter UAVs comprising flight controllers developed in \cite{Arbanas2018}. The follower UAV is equipped with a camera sensor with $85^o$ horizontal FOV and $1280\times 720$ size of image. First, performance of the Kalman filter has been tested by comparing estimation with detection measurements. In case of sudden changes in the target UAV's motion, a lack of detection can be observed while Kalman filter prediction  (Fig.~\ref{fig:xy} and Fig. ~\ref{fig:wh}) gives a continuous information about position and size of the bounding box. Figure \ref{fig:wh} demonstrates another important result of Kalman filter. Namely, during the experiments it has been noticed that relative position of propellers with respect to the body causes volatility of width and height of the bounding box. As it can be seen, the Kalman filter smooths those changes providing steady feedback signal for the distance controller (\ref{eq:th}).

\begin{figure}[ht!]
    \centering
    \vspace*{0.1in}
    \includegraphics[width=0.48\textwidth]{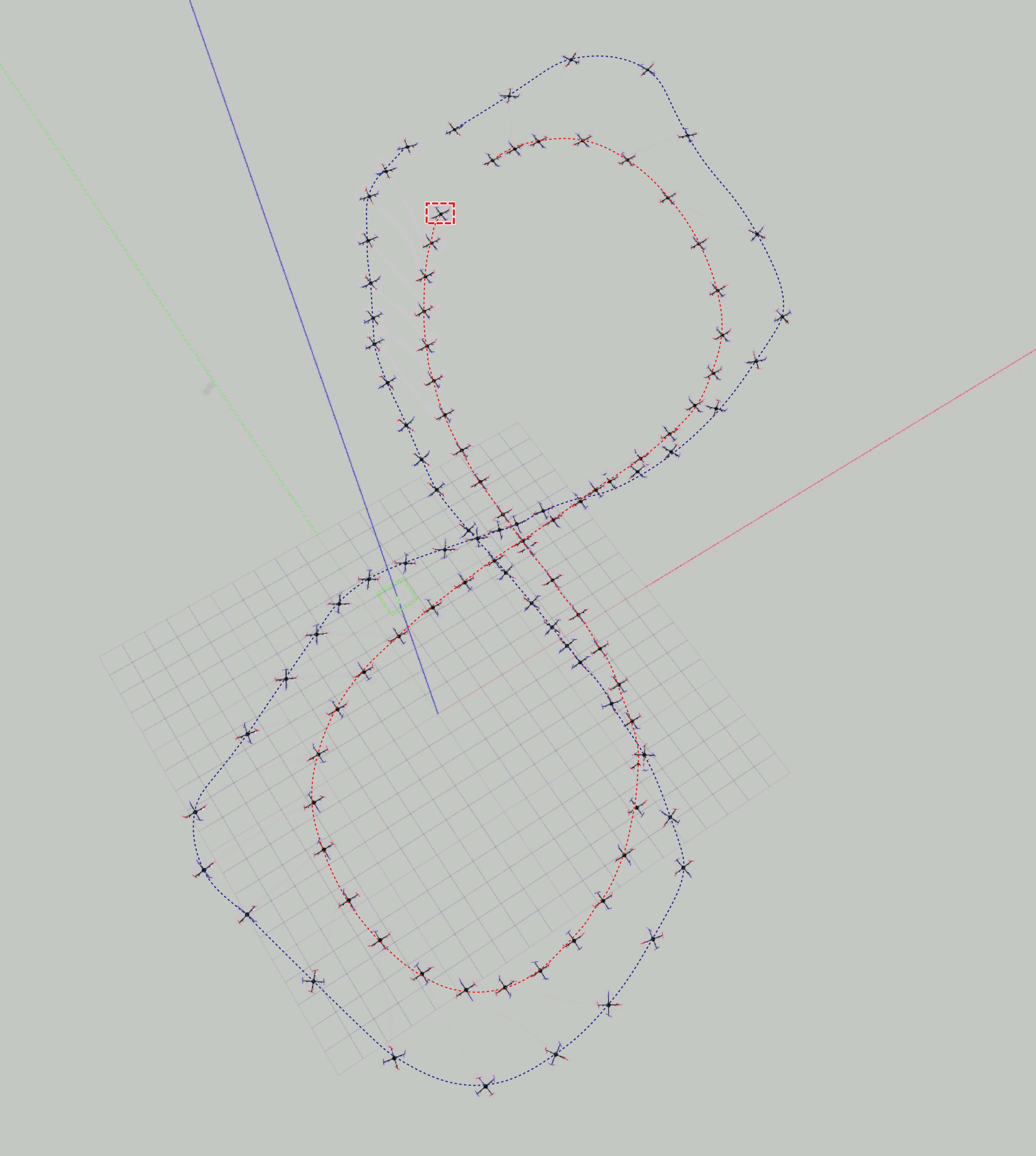}
    \caption{A sequence representation of a successive motion of two UAVs in a final experiment. The follower UAV (blue trajectory) with implemented vision-based system is following a number eight shaped movement of the target UAV (red trajectory). Starting position of the target UAV is marked with a red rectangle.}
    \label{fig:eight}
\end{figure}

Finally, validation of the complete vision-based system for UAV tracking has been done. The target UAV was executing a 3D trajectory in a shape of number eight. Object detection was performed at a speed of 20 FPS. To further confirm the performance of UAV detection, the proposed algorithm was compared to the ground truth at 387 frames taken from the experiment. The trained network achieved 97.88\% mAP score at IOU threshold of 0.5. Moving object tracking with Kalman filter was initiated automatically, on-board the follower UAV, with first confident detection of the target UAV. As Figure \ref{fig:eight} shows, visual servoing successfully keeps the target UAV in the field of view and follows its movement effectively avoiding collisions. A video of conducted experiment is available at \cite{youtube}.

\section{Conclusion}
\label{sec:conclusion}

In this paper, we presented a vision-based system for detection, motion tracking and following of the UAV with a single camera. The position and size of the target UAV is determined by YOLOv3 Tiny model trained on custom dataset. Trained model has capability to detect various UAVs in different environments. By enhancing a detection results with estimation from Kalman filter, steady and reliable visual information has been achieved to be used in visual servoing of the follower UAV. Conducted simulation experiments validate a purpose of developed vision-based system for challenging task of following motion of the target UAV without collisions.

By combining well-known algorithms we have created a powerful real-time system that can be easily modified to execute task of tracking and following of various objects. Proposed solution can be upgraded with ability to avoid static objects and multi-target following. Future work includes testing developed system on a real UAV in outdoor environment where cluttered scene or overlapping objects might be a challenge.

\addtolength{\textheight}{-12cm}   




\section*{Acknowledgment}
This work has been supported by EU H2020 Twinning project AeRoTwin and international robotics challenge MBZIRC 2020.


\balance
\bibliographystyle{ieeetr}
\bibliography{bibliography/bib}

\end{document}